\documentclass[10pt,twocolumn,letterpaper]{article}

\usepackage{iccv}
\usepackage{times}
\usepackage{epsfig}
\usepackage{graphicx}
\usepackage{amsmath}
\usepackage{amssymb}

\newcommand{\ignore}[1]{}

\newcommand{\hide}[1]{}

\usepackage{xcolor}
\usepackage{graphicx}
\usepackage{amsmath}
\usepackage{xspace}
\usepackage{algorithm}
\usepackage{algorithmic}
\usepackage{caption}
\usepackage{csquotes}
\usepackage{url}
\usepackage{breakcites}
\usepackage{tablefootnote}
\usepackage[flushleft]{threeparttable}

\usepackage{booktabs}
\usepackage{enumitem}
\usepackage{multirow, multicol}
\usepackage{mdwlist}
\newcommand{\method}{\textsc{TMI}\xspace}
\newcommand{\fullmethod}{\textsc{Transferability Measurement with Intra-class feature variance}\xspace}

\usepackage[breaklinks=true,bookmarks=false]{hyperref}

\iccvfinalcopy 

\def\iccvPaperID{9680} 

\ificcvfinal\fi

\begin{document}

\title{Fast and Accurate Transferability Measurement by Evaluating Intra-class Feature Variance}

\author{Huiwen Xu\\
Seoul National University \\ Seoul, Republic of Korea\\
{\tt\small xuhuiwen33@snu.ac.kr}
\and
U Kang\\
Seoul National University \\ Seoul, Republic of Korea\\
{\tt\small ukang@snu.ac.kr}
}

\maketitle
\ificcvfinal\thispagestyle{empty}\fi

\begin{abstract}

Given a set of pre-trained models, how can we quickly and accurately find the most useful pre-trained model for a downstream task?
Transferability measurement is to quantify how transferable is a pre-trained model learned on a source task to a target task.
It is used for quickly ranking pre-trained models for a given task and thus becomes a crucial step for transfer learning.
Existing methods measure transferability as the discrimination ability of a source model for a target data before transfer learning, which cannot accurately estimate the fine-tuning performance.
Some of them restrict the application of transferability measurement in selecting the best supervised pre-trained models that have classifiers.
It is important to have a general method for measuring transferability that can be applied in a variety of situations, such as selecting the best self-supervised pre-trained models that do not have classifiers, and selecting the best transferring layer for a target task.

In this work, we propose \method (\fullmethod), a fast and accurate algorithm to measure transferability.
We view transferability as the generalization of a pre-trained model on a target task by measuring intra-class feature variance.
Intra-class variance evaluates the adaptability of the model to a new task, which measures how transferable the model is.
Compared to previous studies that estimate how discriminative the models are, intra-class variance is more accurate than those as it does not require an optimal feature extractor and classifier.
Extensive experiments on real-world datasets show that \method outperforms competitors for selecting the top-5 best models, and exhibits consistently better correlation in 13 out of 17 cases.

\end{abstract}

\section{Introduction}
\label{sec:Intro}

Transfer learning is an important concept in the field of machine learning, revolutionizing the way models are trained and applied to various domains.
Transfer learning has proven to be particularly valuable when labeled data in the target domain are scarce or costly to obtain.
Within the broader scope of transfer learning, several variants have emerged, such as source-free domain adaptation~\cite{dine, iterlnl, tan, jeon}, multi-source transfer learning~\cite{multi-pd, multi-ot, jeon, lee}, heterogeneous transfer learning~\cite{glg, heuda}, and open-set domain adaptation~\cite{open-ad, open-dy}.

\begin{figure}[!t]
	\begin{center}
	\centerline{\includegraphics[width=\columnwidth]{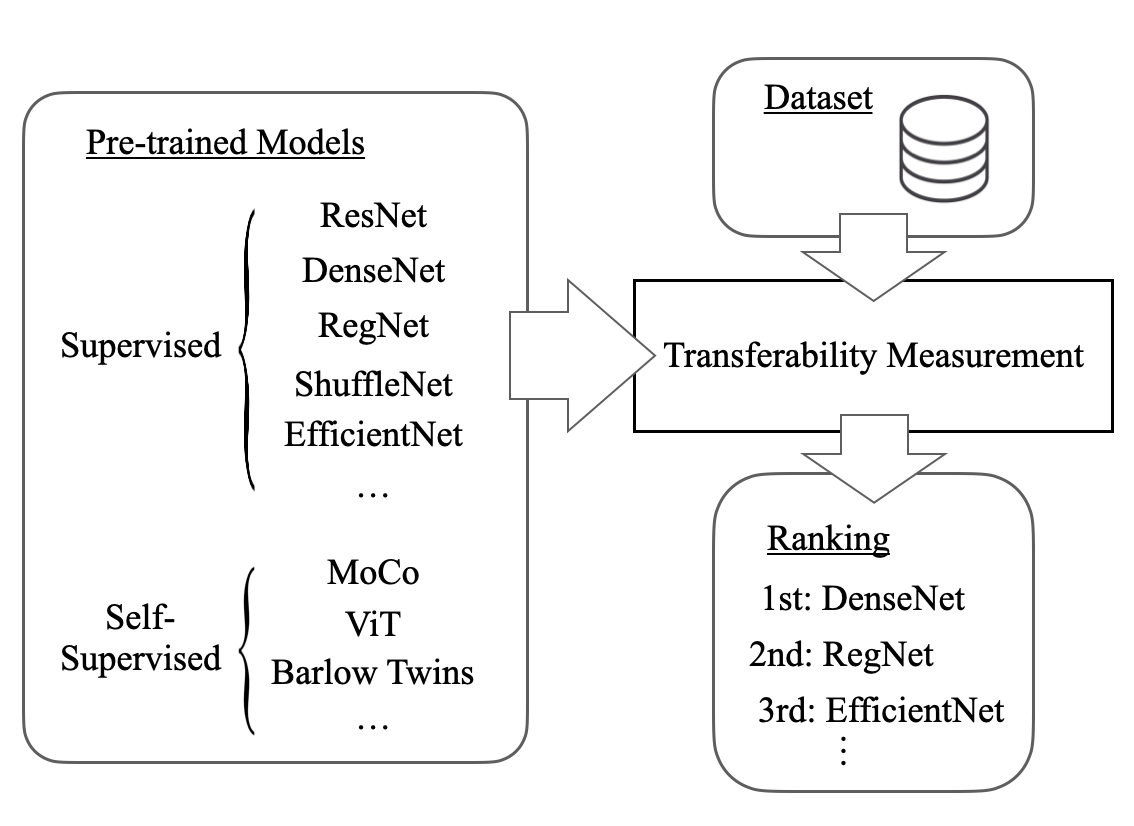}}	
	\caption{
	Illustration of selecting the best pre-trained model for a given target task.
	}
	\label{fig:transferability}
	\end{center}
\end{figure}

In transfer learning, fine-tuning a pre-trained model becomes a simple and effective way to improve the performance of a downstream task.
A crucial step in transfer learning is to quickly and accurately select the most helpful pre-trained model in a set of given pre-trained models.
Transferability measurement is to quantify how transferable is a pre-trained model learned on a source task to a target task.
%
As shown in Figure.~\ref{fig:transferability},
transferability measurement is used for ranking pre-trained models for a given task, and is a crucial step for transfer learning.


A desired method for transferability measurement should be faster than running a transfer learning algorithm, accurately select the best pre-trained model, and be general to be implemented in various common scenarios.
Taskonomy~\cite{taskonomy} and Task2Vec~\cite{task2vec} are computationally expensive due to their learning with data.
NCE~\cite{nce} and LEEP~\cite{leep} are free of training but cannot be applied to general cases, such as selecting self-supervised pre-trained models and selecting the best transferring layer.
LogMe~\cite{logme} and H-Score~\cite{H-score} also cannot be applied to select the best transferring layer.
Instead of measuring transferability, they measure how discriminative a pre-trained model is to a target task before implementing transfer learning, which cannot estimate fine-tuning performance.
TransRate~\cite{transrate} is simple and general to be applied for transferability measurement, but too slow due to the computation of the entropy of whole hidden representations.
%


\begin{figure*}[ht]
    \centering	
    \setlength\tabcolsep{-0.1mm}
	\begin{tabular}{cccccc}
		\vspace{-0.1in}
		\includegraphics[width=0.33\columnwidth]{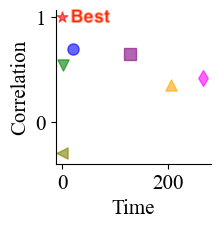}		&
		\includegraphics[width=0.33\columnwidth]{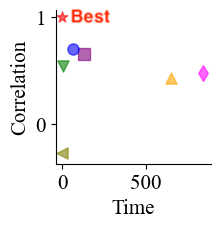}    	&
		\includegraphics[width=0.33\columnwidth]{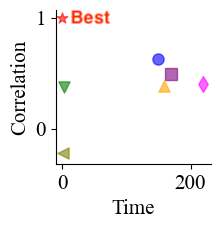}			&
		\includegraphics[width=0.33\columnwidth]{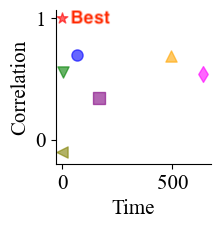}			&
		\includegraphics[width=0.33\columnwidth]{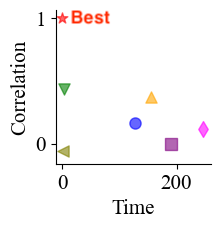}				&
		\includegraphics[width=0.33\columnwidth]{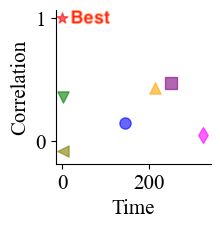}		\\
		\small{(a) Caltech-101}	& \small{(b) Caltech-256}	& \small{(c) CIFAR-10} & \small{(d) CIFAR-100} & \small{(e) MNIST}	& \small{(f) FashionMNIST} \\
		\vspace{-0.1in}
		\includegraphics[width=0.33\columnwidth]{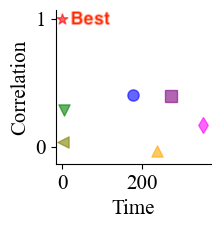}				&
		\includegraphics[width=0.33\columnwidth]{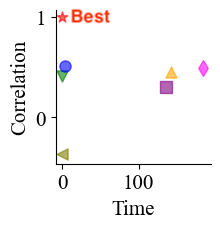}		&
		\includegraphics[width=0.33\columnwidth]{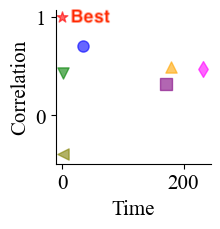}			&
		\includegraphics[width=0.355\columnwidth]{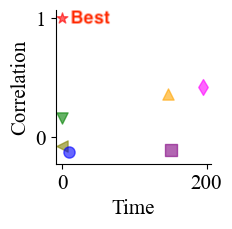}				&
		\includegraphics[width=0.33\columnwidth]{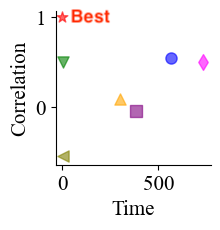}				&
		\includegraphics[width=0.33\columnwidth]{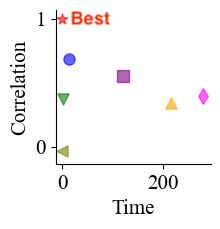}		\\
		\small{(g) SVHN}	& \small{(h)FlowerPhotos}	&	\small{(i) EuroSAT}	&	\small{(j) Chest X-Ray}	&	\small{(k) VisDA}	&	\small{(l) FGVC-Aircraft}	\\
		\vspace{-0.1in}

		\includegraphics[width=0.33\columnwidth]{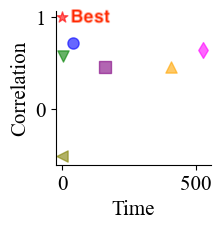}			&
		\includegraphics[width=0.33\columnwidth]{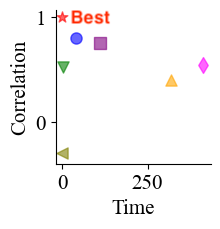}				&
		\includegraphics[width=0.33\columnwidth]{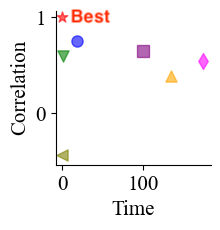}				&
		\includegraphics[width=0.33\columnwidth]{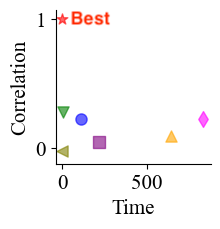}			&
		\includegraphics[width=0.33\columnwidth]{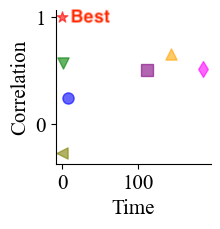}		&
		\includegraphics[width=0.35\columnwidth]{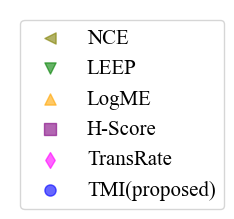}				\\		
		\small{(m) CUB-200}	& \small{(n) Cars} &	 \small{(o) DTD} & \small{(p) Food-101}	& \small{(q) Oxford-IIIT Pet}	&	\\
	\end{tabular}
	
	\caption{
	Total running time (seconds) vs. Kendall's Correlation between transferability scores and test accuracies
	for selecting the best model among fifty pre-trained models.
	\method provides the best trade-off between running time and correlation. 
	}
	\label{fig:time_correlation}
\end{figure*}

In this paper, we propose \method (\fullmethod), a simple but effective method for transferability measurement.
We deem transferability as the generalization of a pre-trained model on a target task.
To quantify the degree of generalization, we measure the intra-class variance by the conditional entropy of target representation given a label.
Intra-class variance evaluates the adaptability of the model to a new task, which measures how transferable the model is.
Compared to previous studies that estimate how discriminative the models are, intra-class variance is more accurate than those as it does not require an optimal feature extractor and classifier.
Large intra-class variance implies the feature is efficient for learning tasks.
\method can be used for selecting not only the best supervised model but also the best self-supervised models and the best transferring layer.

Our contributions are summarized as follows:

\begin{itemize*} 
	\item \textbf{Algorithm.}
	We propose \method, a fast, accurate, and general method for measuring transferability.
	We view transferability as generalization ability of a pre-trained model on a target task.
	The model generalization is measured by intra-class feature variance.
	\item \textbf{Experiments.}
	We conduct extensive experiments on seventeen datasets with fifty supervised pre-trained models and eleven self-supervised pre-trained models.
	\method outperforms competitors for selecting the top-5 pre-trained architecture in 14 out of 17 cases and selecting the best source data in all cases.
    \method also shows the best trade-off of the running time and transferability, outperforming competitors in 12 out of 17 cases (see Figure.~\ref{fig:time_correlation}).
	\item \textbf{Case study.}
	We evaluate the transferability in more general cases, such as applying to self-supervised pre-trained models and selecting the best transferring layer.
	Compared to competitors, \method achieves the highest correlation coefficient in 13 out of 17 for selecting the best self-supervised pre-trained models, and in 16 out of 17 datasets for selecting the best transferring layer.
	
\end{itemize*}

The rest of this paper is organized as follows.
We introduce related works for transferability measurement in Section~\ref{sec:Related},
and propose our method \method in Section~\ref{sec:Methods}.
After presenting experimental results in Section~\ref{sec:Experiments},
we conclude in Section~\ref{sec:Conclusion}.

\section{Related Works}
\label{sec:Related}
%
%

We assume a pre-trained model $\theta=(w, h)$ in a source domain where $w$ is a feature extractor, and $h$ is a classifier (or head).
We transfer the model $\theta$ to a target domain and add a new classifier $k$ at the end of the feature extractor $w$ before fine-tuning.
Since the target task is different from the source task in most cases, the pre-trained classifier $h$ is inappropriate to solve the target task.
Thus, we fine-tune the model by minimizing the cross entropy loss on the target dataset $D=\{(x_i,y_i)\}^n_{i=1}$:

\begin{equation}
	\setlength{\abovedisplayshortskip}{-2pt}
	w^*, k^* = \underset{w, k}{\arg\min} \mathcal{L} (w, k)
	\setlength{\belowdisplayskip}{-2pt}
\end{equation}

\noindent where a feature extractor $w$ maps an input $x_i$ to a representation $r_i = w(x_i)$,
and a classifier $h$ maps the representation $r_i$ to a probability distribution $\hat{y}_i = \sigma (h(r_i))$ where $\sigma(\cdot)$ is an activation function.
Note that $\hat{y}_{i}$ is a vector of length $n_c$, where $n_c$ is the number of classes,
and
$\mathcal{L(\cdot)}$ is the cross entropy loss, which is equivalent to the average negative log-likelihood:

\begin{equation}
	\mathcal{L} (w, k) = - \frac{1}{n}\sum^{n}_{i=1}\log P( \hat{y}_{i, c} | x_i ; w, k)
\end{equation}

\noindent where $\hat{y}_{i,c}$ is the probability of $i$-th input instance belonging to its ground truth.

The transferability of a source task $S$ to a target task $T$ is defined by the log-likelihood~\cite{nce}:
\begin{equation}
	Trf_{(S\rightarrow T)} = \frac{1}{n}\sum^{n}_{i=1}\log P( \hat{y}_{i, c} | x_i ; w^*, k^*)
	\label{eq:opt_trf}
\end{equation}

Taskonomy~\cite{taskonomy} and Task2Vec~\cite{task2vec} train the source model using target data to obtain the optimal feature extractor $w^*$ and the optimal classifier $k^*$.
These methods are inconvenient to measure the transferability due to their training with data.
There have been works focusing on estimating a bound on the log-likelihood to evaluate the transferability without training.
NCE~\cite{nce} proposes negative conditional entropy, which is drawn from the empirical joint distribution over ground truths and predicted labels from a pre-trained model.
Similarly, LEEP~\cite{leep} replaces the ground truth with the average log-likelihood to evaluate the transferability.
NCE and LEEP are fast, but are restricted to supervised pre-trained models because the algorithms rely on the outputs of the pre-trained classifier.
To apply measurements in more general cases, LogME~\cite{logme} calculates the logarithm of maximum evidence given extracted features, and H-Score~\cite{H-score} solves an HGR maximum correlation problem.
LogMe~\cite{logme} and H-Score~\cite{H-score} still cannot be applied to select the best transferred layer since they extract features from only the ultimate layers of the pre-trained feature extractors.
TransRate~\cite{transrate} is the most recent algorithm and applicable to self-supervised models and transferring layer selection.
However, it is time-consuming since it requires to calculate mutual information between extracted feature and ground truth.
To simplify the problem, \cite{nce, leep, logme, H-score, transrate} assume that the optimal model can be obtained by re-training the classifier while freezing the feature extractor.
The assumption goes against the purpose of transferability measurement because we re-train both feature extractor and classifier in general transferring methods. 
The performance of re-training only the classifier is even worse than training a target model from scratch~\cite{how_transferable}.
In this paper, we propose a fast, accurate, and general transferability measurement method that predicts fine-tuned performance of a pre-trained model to a new target task.
%
%

\section{Proposed Method}
\label{sec:Methods}
We propose \method to measure the transferability between a pre-trained model and a target task.

\subsection{Overview}
\label{subsec:overview}
Our goal is to measure the transferability in a fast, accurate, and general manner.
The following challenges need to be addressed for the goal.


\begin{enumerate}[nosep, label=\textbf{C\arabic*}]
	\item \textbf{Speed.}
	Training a pre-trained model on the target data requires a high computational cost. 	
	How can we measure the transferability without training on the target data?
	\item \textbf{Accuracy.}
	Existing methods evaluate the transferability that estimate the performance of the optimal target model obtained by re-training only the classifier, not the feature extractor, which is uncommon in transfer learning.
	How can we accurately measure the transferability for predicting the fine-tuning performance?
	\item \textbf{Generality.}
	How can we measure the transferability such that it is applicable to general settings, including selecting the best supervised pre-trained model, the best self-supervised pre-trained models, and the transferring layer?
\end{enumerate}

We address the aforementioned challenges with the following main ideas:

\begin{enumerate}[nosep, label=\textbf{I\arabic*}]
	\item
	We assess the transferability by evaluating the compactness and sparseness of target representations with only a one-way forward propagation.
	\item
	We relax the assumption proposed in existing methods that the pre-trained feature extractor can extract all the information related to the target task.
	We view transferability as the generalization ability of a pre-trained model on a target task.
	We estimate the generalization ability using intra-class variance, which implies how efficient the feature is for learning tasks.
	\item
	We measure the transferability using intra-class variance of latent representations, which is applicable to various settings including self-supervised pre-trained models and transferring layer selection.
\end{enumerate}

\subsection{\method}
%
Given a pre-trained source model and a target dataset with labels, our goal is to measure the transferability between the source model and the target task to predict the fine-tuning performance.
Existing methods measure the performance of re-trained classifier while freezing the feature extractor, which is time-consuming. 
To address the issues, our proposed \method computes the transferability as the intra-class variance of a pre-trained model on a target task.
Intra-class variance evaluates the adaptability of the model to a new task, which measures how transferable the model is.
It is unnecessary for \method to obtain optimal classifier which is a time-consuming process.

\noindent \textbf{\method}
	The transferability of a pre-trained feature extractor $\omega_s$ from a source task $S$ to a target task $T$, denoted by $Trf_{(S\rightarrow T)}$, is measured by \method with the intra-class feature variance obtained from the pre-trained model $\omega_s$.

\begin{equation}
	\label{eq:method}
	Trf_{(S\rightarrow T)} := H(\omega_s(X)|Y)=\sum_{c=1}^C \frac{n_c}{n} H(\omega_s(X_c))
\end{equation}

\noindent where $X$ and $Y$ represent the target feature and the target label, respectively.
$C$ represents the number of classes and $X_c=\{x|Y=c\}$ is the set of target features in the $c$-th class.
$n$ and $n_c$ represent the number of total instances and number of instances in the $c$-th class, respectively.
We measure the intra-class variance of target representations using conditional entropy.
The intra-class variance encourages the pre-trained model to have a generalization ability for learning tasks.
%
%

\noindent \textbf{How to utilize \method?}
Given a set of pre-trained models and a target dataset with labels, we first obtain target representations of pre-trained models by one-way forward propagation.
Next, we compute the transferability score using Equation~\eqref{eq:method}.
We select the pre-trained model with the highest transferability score and then fine-tune the model using the target data. 
We argue that the larger the TMI, the higher the transfer performance. 

\section{Experiments}
\label{sec:Experiments}
%
We present experimental results to answer the following questions about \method:

\begin{enumerate}[nosep, label=\textbf{Q\arabic*}]
	\item \textbf{Evaluation of transferability estimation} (Section~\ref{subsec:exp_supervised}). Does \method accurately select the best pre-trained model compared to baselines? Is the transferability score measured by \method highly correlated to the transfer performance?
	\item \textbf{Comparison of ICV measures} (Section~\ref{subsec:exp_measure}). What is the best measurement to evaluate Intra-Class Variance (ICV)?
	\item \textbf{Evaluation of generalization ability} (Section~\ref{subsec:exp_dis}). Does intra-class variance estimate generalization ability of a pre-trained model on a target task?
	\item \textbf{General applications} (Section~\ref{subsec:exp_application}). Does \method accurately estimate the transferability in general cases, such as selecting the best self-supervised pre-trained model and the best transferring layer?
	\item \textbf{Hyperparameter sensitivity} (Section~\ref{subsec:exp_hyp}). How sensitive is \method to hyperparameters?
\end{enumerate}

\subsection{Experimental Setup}
\label{subsec:setup}
We present pre-trained models, datasets, baselines, evaluation, estimator, and hyperparameters for our experiments.

\paragraph{Pre-trained models.}
We use supervised model groups offered by Pytorch:
ResNet, DenseNet, MnasNet, ShuffleNet V2, MobileNet V2, MobileNet V3, ResNeXt, Wide ResNet, EfficientNet, EfficientNet V2, and RegNet.
We also utilize self-supervised model groups, which have only feature extractors without classifiers.
We obtain Swin Transformer and Vision Transformer from Pytorch;
we get
MoCo v1, MoCo v2, Barlow Twins, SwAV, SimSiam, and Dino from Gitlab repositories.
Note that we select fifty supervised pre-trained models and eleven self-supervised pre-trained models from the aforementioned model groups.
%

\paragraph{Datasets.}
We use seventeen image datasets summarized in Table~\ref{table: datasets}.
\hide{
Caltech-101\footnote{\url{https://data.caltech.edu/records/mzrjq-6wc02}}~\cite{caltech},
Caltech-256\footnote{\url{https://data.caltech.edu/records/nyy15-4j048}}~\cite{caltech},
CIFAR-10\footnote{\url{https://www.cs.toronto.edu/~kriz/cifar.html}}~\cite{cifar},
CIFAR-100\footnotemark[3]~\cite{cifar},
MNIST\footnote{\url{http://yann.lecun.com/exdb/mnist/}}~\cite{mnist},
FashionMNIST\footnote{\url{https://github.com/zalandoresearch/fashion-mnist}}~\cite{fashionmnist},
SVHN\footnote{\url{http://ufldl.stanford.edu/housenumbers/}}~\cite{svhn},
FlowerPhotos\footnote{\url{https://www.tensorflow.org/datasets/catalog/tf_flowers}}, 
EuroSAT\footnote{\url{https://github.com/phelber/eurosat}}~\cite{eurosat},
Chest X-Ray\footnote{\url{https://data.mendeley.com/datasets/rscbjbr9sj/2}}~\cite{chest},
VisDA\footnote{\url{http://ai.bu.edu/visda-2017/\#browse}}~\cite{visda},
FGVC-Aircraft\footnote{\url{https://www.robots.ox.ac.uk/~vgg/data/fgvc-aircraft/}}~\cite{aircraft},
CUB-200\footnote{\url{http://www.vision.caltech.edu/datasets/cub_200_2011/}}~\cite{cub},
Cars~\footnote{\url{https://ai.stanford.edu/~jkrause/cars/car_dataset.html}}~\cite{cars},
DTD\footnote{\url{https://www.robots.ox.ac.uk/~vgg/data/dtd/}}~\cite{dtd},
Food-101\footnote{\url{https://data.vision.ee.ethz.ch/cvl/datasets_extra/food-101/}}~\cite{food},
and
Oxford-IIIT Pet\footnote{\url{https://www.robots.ox.ac.uk/~vgg/data/pets/}}~\cite{pet}
in experiments, which are summarized in Table~\ref{table: datasets}.
Note that each dataset represents a classification task.
}
We randomly split each dataset into a train set and test set by 8:2 ratio, except for datasets that already have been separated.

\begin{table}[!h!t]
	\centering
	\caption{Summary of datasets.}
	\label{table: datasets}
	\scalebox{0.7}{
	\begin{threeparttable}
		\begin{tabular}{lrrrr}
			\toprule
			\textbf{Dataset}			&	\textbf{\# of instances (train / test)} &	\textbf{\# of classes}	&		\textbf{Category}\\
			\midrule
			Caltech-101\tnote{1}		&	7,315 / 1,829		&	101		& Multiple domains\\
			Caltech-256\tnote{2}		&	24,485 / 6,122		&	256		& Multiple domains\\
			CIFAR-10\tnote{3}		&	50,000 / 10,000	    &	10		& Multiple domains\\
			CIFAR-100\tnote{3}		&	50,000 / 10,000		&	100		& Multiple domains\\
			MNIST\tnote{4}			&	60,000 / 10,000		&	10		& Digit\\
			FashionMNIST	\tnote{5}	&	60,000 / 10,000		&	10		& Fashion\\
			SVHN\tnote{6}			&	73,257 / 26,032		&	10		& Digit\\
			FlowerPhotos	\tnote{7}	&	2,936 / 734			&	5		& Flower\\
			EuroSAT\tnote{8} 		&	21,600 / 5,400		&	10		& Landscape\\
			Chest X-Ray\tnote{9}	 	&	5,216 / 623			&	2		& Medicine\\
			VisDA\tnote{10} 			&	152,397 / 72,372		&	212		& Multiple domains\\
			FGVC-Aircraft\tnote{11}	&	3,334 / 3,333		&	100		& Aircraft\\
			CUB-200\tnote{12} 		&	9,430 / 2,358		&	200		& Animal\\
			Cars\tnote{13} 			&	8,144 / 8,041		&	196		& Auto\\
			DTD\tnote{14} 			&	4,512 / 1,128 		&	47		& Texture\\
			Food-101\tnote{15} 		&	80,800 / 20,200 		&	101 	& Food\\
			Oxford-IIIT Pet\tnote{16} 	&	3,680 / 3,669		&	37		& Animal\\
			\bottomrule
		\end{tabular}

		\begin{tablenotes}
		    \item[1] \url{https://data.caltech.edu/records/mzrjq-6wc02}
		    \item[2] \url{https://data.caltech.edu/records/nyy15-4j048}
		    \item[3] \url{https://www.cs.toronto.edu/~kriz/cifar.html}
		    \item[4] \url{http://yann.lecun.com/exdb/mnist/}
		    \item[5] \url{https://github.com/zalandoresearch/fashion-mnist}
		    \item[6] \url{http://ufldl.stanford.edu/housenumbers/}
		    \item[7] \url{https://www.tensorflow.org/datasets/catalog/tf_flowers}
		    \item[8] \url{https://github.com/phelber/eurosat}
		    \item[9] \url{https://data.mendeley.com/datasets/rscbjbr9sj/2}
		    \item[10] \url{http://ai.bu.edu/visda-2017/\#browse}
		    \item[11] \url{https://www.robots.ox.ac.uk/~vgg/data/fgvc-aircraft/}
		    \item[12] \url{http://www.vision.caltech.edu/datasets/cub_200_2011/}
		    \item[13] \url{https://ai.stanford.edu/~jkrause/cars/car_dataset.html}
		    \item[14] \url{https://www.robots.ox.ac.uk/~vgg/data/dtd/}
		    \item[15] \url{https://data.vision.ee.ethz.ch/cvl/datasets_extra/food-101/}
		    \item[16] \url{https://www.robots.ox.ac.uk/~vgg/data/pets/}		    		
	    \end{tablenotes}
	\end{threeparttable}
	}
\end{table}

\paragraph{Baselines.}
We compare the proposed method with the following baselines.
%
NCE~\cite{nce} defines the transferability as negative conditional entropy between ground truths and predicted labels from a pre-trained model.
LEEP~\cite{leep} measures negative conditional entropy between ground truths and predicted average log-likelihood.
H-Score~\cite{H-score} estimates the transferability by solving an HGR maximum correlation problem.
LogME~\cite{logme} proposes the logarithm of maximum evidence given extracted features.
TransRate~\cite{transrate} measures the transferability by the mutual information between target labels and target representations.
We omit the comparison with Taskonomy~\cite{taskonomy} and Task2Vec~\cite{task2vec} due to tremendous time costs.

\paragraph{Evaluation.}
We evaluate the performance of transferability measurement methods in two ways.
First, do they accurately predict the best model?
Second, are the transferability scores correlated to fine-tuned accuracy?
We adopt Kendall correlation coefficient to measure the correlations between the transferability and fine-tuned accuracy.
Kendall correlation coefficient tests the ordinal association between two measured quantities.
Kendall correlation coefficient has values between -1 and 1, and correlations between two variables are high when the values are close to 1.

\paragraph{Estimator.}
We adopt $k$-nearest estimator~\cite{entropy} to evaluate the continuous entropy in Equation~\eqref{eq:method}.
Compared to kernel density estimators~\cite{kernel1, kernel2}, $k$-nearest estimator vastly reduces errors by sacrificing time, especially when data are too big.
\method greatly reduces the size of data to be calculated at once because it computes conditional entropy of representation given label.

\paragraph{Hyperparameters.}
We use greedy searching algorithm to find the best hyperparameters for fine-tuning.
For transferability measurement, the only hyperparameter in our proposed \method is the number $n_k$ of neighbors, which is used in the entropy estimator~\cite{entropy}.
We discuss the hyperparameter sensitivity to transferability in Section~\ref{subsec:exp_hyp}.

\subsection{Evaluation of Transferability  Estimation}
\label{subsec:exp_supervised}

Remind that our goal is to quickly and accurately select the best pre-trained model among a pool of well-trained models.
A pre-trained model is determined by two components, source data and model architecture.
Thus, we conduct experiments to select the best model architecture and the best source data. 

\paragraph{Selecting the best model architecture.}
We evaluate how the transferability measurement methods accurately select the best source model for a target model.
Table~\ref{table:supervised_top5} shows the ratio of targets tasks
where their top-5 best source models include the model estimated by each transferability measurement method.
Note that \method gives the best ratio,
accurately selecting the best pre-trained models in 14 out of 17 cases,
which is much better than the ratio 9 / 17 of the second-best model TransRate.

\begin{table}[t]
	\caption
	{
	Ratio of target tasks where their top-5 best source models include the model estimated by each transferability measurement method. The best result is in bold.
Note that \method gives the best ratio,
accurately selecting the best pre-trained models in 14 out of 17 cases.
	}
	\centering
	\label{table:supervised_top5}
	\scalebox{0.75}{
	\begin{tabular}{lcccccc}
	\toprule
	\textbf{Methods}       	& \textbf{NCE}   & \textbf{LEEP}      & \textbf{LogMe}		& \textbf{H-Score} & \textbf{TransRate} & \textbf{\method}  \\
	\midrule
	Top-5 & 3 / 17 & 5 / 17 & 4 / 17 & 8 / 17 & 9 / 17 & \textbf{14 / 17} \\
	\bottomrule
	\end{tabular}
	}
\end{table}

We also compare the correlations between fine-tuned accuracies and transferability scores in Table~\ref{table:supervised}.
Note that our proposed \method shows the best Kendall correlation in 13 out of 17 datasets, outperforming all the competitors.
%
Some baselines even give negative coefficients, which means that their predictions are far from the correct ranking.


\begin{table*}[!t]
	\caption
	{
	Correlations between fine-tuned accuracies and transferability scores when transferring supervised pre-trained models trained on ImageNet.
The best result is in bold.
Our proposed \method gives the best Kendall correlation in 13 out of 17 cases, outperforming competitors.
	}
	\centering
	\label{table:supervised}
	\begin{tabular}{lcccccc}
		\toprule
        \multirow{2}{*}{\textbf{Dataset}}	& \multicolumn{6}{c}{\textbf{Kendall Correlation Coefficient}}	\\
                & \textbf{NCE}   & \textbf{LEEP}      & \textbf{LogMe}		& \textbf{H-Score} & \textbf{TransRate} & \textbf{\method}         \\
        \midrule
Caltech-101     & 0.429          & 0.425          & 0.316  & 0.362 & 0.474 & \textbf{0.524} \\
Caltech-256     & \textbf{0.686} & 0.455          & 0.492  & 0.503 & 0.509 & 0.481          \\
CIFAR-10        & -0.140         & 0.324          & 0.275  & 0.323 & 0.364 & \textbf{0.527} \\
CIFAR-100       & -0.202         & 0.340          & 0.351  & 0.398 & 0.432 & \textbf{0.497} \\
MNIST           & 0.057          & \textbf{0.282} & 0.083  & 0.137 & 0.038 & 0.137          \\
FashionMNIST    & -0.328         & 0.127          & -0.038 & 0.121 & 0.125 & \textbf{0.467} \\
SVHN            & -0.023         & 0.091          & -0.002 & 0.092 & 0.206 & \textbf{0.394} \\
FlowerPhotos    & 0.117          & 0.139          & 0.224  & 0.321 & 0.392 & \textbf{0.499} \\
EuroSAT         & \textbf{0.269} & 0.082          & -0.005 & 0.102 & 0.051 & 0.085          \\
Chest X-Ray     & 0.071          & 0.152          & 0.171  & 0.333 & 0.285 & \textbf{0.395} \\
VisDA           & 0.298          & 0.240          & 0.329  & 0.334 & 0.304 & \textbf{0.341} \\
FGVC-Aircraft   & 0.366          & -0.011         & 0.216  & 0.498 & 0.421 & \textbf{0.500} \\
CUB-200         & 0.415          & 0.098          & 0.325  & 0.410 & 0.519 & \textbf{0.523} \\
Cars            & 0.270          & 0.506          & 0.175  & 0.432 & 0.455 & \textbf{0.521} \\
DTD             & -0.074         & \textbf{0.501} & 0.479  & 0.366 & 0.499 & 0.401          \\
Food-101        & 0.261          & 0.515          & 0.395  & 0.421 & 0.448 & \textbf{0.537} \\
Oxford-IIIT Pet & 0.536          & 0.527          & 0.418  & 0.493 & 0.500 & \textbf{0.576} \\
		\bottomrule
	\end{tabular}
\end{table*}

\paragraph{Selecting the best source data.}
We fine-tune eighteen pre-trained models with the same model architecture ResNet-50, but different source datasets, including ImageNet and seventeen datasets introduced in Section~\ref{subsec:setup}.
We predict the best source data and compare correlations with the ground truths.
Table~\ref{table:source_best} shows that our proposed \method
accurately predicts the best source data in all the 17 cases, outperforming all the competitors.
Our proposed \method gives the best Kendall correlation consistently in 12 out of 17 cases, as shown in Table~\ref{table:source}.


\begin{table}[!t]
	\caption
	{
	Ratio of target tasks where the transferability measurement methods accurately find the best source data.
The best method is in bold.
	\method predicts the best source data in all cases, outperforming competitors.
	}
	\centering
	\label{table:source_best}
	\scalebox{0.75}{
	\begin{tabular}{lcccccc}
	\toprule
	\textbf{Methods}       	& \textbf{NCE}   & \textbf{LEEP}      & \textbf{LogMe}		& \textbf{H-Score} & \textbf{TransRate} & \textbf{\method}  \\
	\midrule
	Top-1  & 1 / 17  & 2 / 17    & 3 / 17       & 2 / 17     & 13 / 17        & \textbf{17 / 17} \\
	\bottomrule
	\end{tabular}
	}
\end{table}

\begin{table*}[!t]
	\caption
	{
	Correlations between fine-tuned accuracies and transferability scores when transferring ResNet-50 models trained on different source data.
	The best result is in bold.
	\method shows the best Kendall correlation in 12 out of 17 cases, outperforming all the competitors.
	}
	\centering
	\label{table:source}
	\begin{tabular}{lcccccc}
		\toprule
        \multirow{2}{*}{\textbf{Dataset}}	& \multicolumn{6}{c}{\textbf{Kendall Correlation Coefficient}}	\\
                & \textbf{NCE}   & \textbf{LEEP}      & \textbf{LogMe}		& \textbf{H-Score} & \textbf{TransRate} & \textbf{\method}  \\
        \midrule
Caltech-101     & -0.294 & 0.542          & 0.353          & 0.647          & 0.425          & \textbf{0.699} \\
Caltech-256     & -0.268 & 0.542          & 0.437          & 0.660          & 0.477          & \textbf{0.699} \\
CIFAR-10        & -0.216 & 0.373          & 0.383          & 0.490          & 0.399          & \textbf{0.630} \\
CIFAR-100       & -0.098 & 0.556          & 0.686          & 0.345          & 0.542          & \textbf{0.699} \\
MNIST           & -0.061 & \textbf{0.439} & 0.372          & -0.007         & 0.115          & 0.162          \\
FashionMNIST    & -0.086 & 0.362          & 0.430          & \textbf{0.475} & 0.046          & 0.150          \\
SVHN            & 0.039  & 0.285          & -0.033         & 0.393          & 0.170          & \textbf{0.407} \\
FlowerPhotos    & -0.359 & 0.412          & 0.456          & 0.307          & 0.490          & \textbf{0.516} \\
EuroSAT         & -0.388 & 0.428          & 0.490          & 0.322          & 0.467          & \textbf{0.704} \\
Chest X-Ray     & -0.072 & 0.164          & 0.359          & -0.111         & \textbf{0.425} & -0.123         \\
VisDA           & -0.538 & 0.498          & 0.094          & -0.039         & 0.498          & \textbf{0.546} \\
FGVC-Aircraft   & -0.033 & 0.373          & 0.339          & 0.556          & 0.399          & \textbf{0.688} \\
CUB-200         & -0.503 & 0.582          & 0.455          & 0.464          & 0.647          & \textbf{0.723} \\
Cars            & -0.294 & 0.529          & 0.403          & 0.752          & 0.542          & \textbf{0.804} \\
DTD             & -0.438 & 0.595          & 0.393          & 0.647          & 0.542          & \textbf{0.757} \\
Food-101        & -0.020 & \textbf{0.281} & 0.092          & 0.046          & 0.229          & 0.225          \\
Oxford-IIIT Pet & -0.268 & 0.569          & \textbf{0.660} & 0.503          & 0.516          & 0.248      	  \\
		\bottomrule
	\end{tabular}
\end{table*}

\subsection{Comparison of ICV Measures}
\label{subsec:exp_measure}

Intra-Class Variance (ICV) is a general concept widely used in metric learning.
In addition to the conditional entropy introduced in Section~\ref{sec:Methods}, we additionally choose four metric learning methods, Contrast~\cite{contrast}, Center~\cite{center}, SNCA~\cite{snca}, and MS~\cite{ms}.
We use the intra-class variance parts of their objective functions to measure the transferability.
We apply the five aforementioned methods to measure the correlations between fine-tuned accuracies and transferability scores.
Conditional Entropy (CE) shows the best correlations than the other four metric learning measurements in 15 out of 17 datasets, outperforming all the competitors, as shown in Table~\ref{table:measure}.
M. Boudiaf et al.~\cite{pce} demonstrate that the conditional entropy is a lower bound of intra-class variance of the four metric learning methods.

\begin{table}[!t]
	\caption
	{
	Comparison of intra-class variance measurements. Conditional Entropy (CE) shows the best correlations consistently in 15 out of 17 cases.
	The best result is in bold.
	}
	\centering
	\label{table:measure}
	\scalebox{0.8}{
	\begin{tabular}{lccccc}
		\toprule
		\multirow{2}{*}{\textbf{Dataset}} 		& \multicolumn{5}{c}{\textbf{Kendall Correlation Coefficient}}	\\
		& 	\textbf{Contrast}		&	\textbf{Center}		&	\textbf{SNCA}			&	\textbf{MS}		&	\textbf{CE}	\\
		\midrule
		Caltech-101     & 0.274 & 0.283 & 0.380 & 0.358 & \textbf{0.442}\\
		Caltech-256     & 0.295 & 0.347 & 0.261 & 0.347 & \textbf{0.354}\\
		CIFAR-10        & 0.328 & 0.322 & 0.275 & 0.299 & \textbf{0.369}\\
		CIFAR-100       & 0.374 & 0.282 & 0.207 & 0.200 & \textbf{0.376}\\
		MNIST           & \textbf{0.383} & 0.267 & 0.306 & 0.228 & 0.374\\
		FashionMNIST    & 0.332 & 0.225 & 0.260 & 0.337 & \textbf{0.374}\\
		SVHN            & 0.303 & 0.359 & 0.369 & 0.360 & \textbf{0.383}\\
		FlowerPhotos    & 0.209 & 0.263 & 0.291 & 0.277 & \textbf{0.372}\\
		EuroSAT         & 0.292 & 0.237 & 0.367 & 0.268 & \textbf{0.400}\\
		Chest X-Ray     & 0.292 & 0.289 & 0.210 & 0.232 & \textbf{0.357}\\
		VisDA           & 0.259 & \textbf{0.352} & 0.200 & 0.329 & \textbf{0.352}\\
		FGVC-Aircraft   & 0.220 & 0.322 & 0.244 & 0.315 & \textbf{0.351}\\
		CUB-200         & 0.368 & 0.252 & 0.379 & 0.385 & \textbf{0.431}\\
		Cars            & 0.351 & 0.285 & \textbf{0.398} & 0.268 & 0.248\\
		DTD             & 0.236 & 0.322 & 0.228 & 0.206 & \textbf{0.327}\\
		Food-101        & 0.322 & 0.267 & 0.246 & 0.269 & \textbf{0.484}\\
		Oxford-IIIT Pet & 0.301 & 0.202 & 0.363 & 0.274 & \textbf{0.449}\\
%
		\bottomrule
	\end{tabular}
	}
\end{table}

\subsection{Evaluation of Generalization Ability}
\label{subsec:exp_dis}
We perform experiments on two control groups to demonstrate that transferability is relevant to the generalization ability of a pre-trained model on a target task.
In Group A, we use a pre-trained ResNet-50 model trained on ImageNet and fine-tune it using the dataset $D$.
In Group B, we train a ResNet-50 model trained from scratch on dataset $D$ and fine-tune it on the same dataset.
Dataset $D$ is one of the datasets introduced in Section~\ref{subsec:setup}.

Table~\ref{table:dis} shows the difference of fine-tuning performance and that of transferability between two groups.
\enquote{+} represents the difference is larger than 0 while \enquote{-} represents that is less than 0.
Group A has higher transfer accuracies than Group B in most cases
since models in Group A have learned a large amount of information from the ImageNet, which can potentially benefit downstream tasks.

Our proposed \method and TransRate have the same symbols for transfer accuracy in 14 out of 17 cases, while the other four baselines do not provide similar symbols.
TransRate also estimates well, but it is computationally expensive since it measures the mutual information between target representation and label.
In contrast, \method is more efficient as it considers only intra-class features.


\begin{table}[!t!]
	\centering
	\setlength{\tabcolsep}{1.9pt}
	\caption{Comparison of the consistency of fine-tuning accuracy and transferability.
	\enquote{+} represents the difference is larger than 0 while \enquote{-} represents the difference is less than 0.
	\method has consistent signs with fine-tuning accuracy in 14 out of 17 cases, outperforming its competitors.
}
	\label{table:dis}
	\scalebox{0.75}{
	\begin{tabular}{lccccccc}
		\toprule
		\multicolumn{1}{c}{\multirow{2}{*}{\textbf{Dataset}}} & \textbf{Accuracy} & \multicolumn{6}{c}{\textbf{Transferability (A - B)}}                          \\
		\multicolumn{1}{c}{}                 &   \textbf{(A - B)}        & \multicolumn{1}{l}{\textbf{NCE}} & \multicolumn{1}{l}{\textbf{LEEP}} & \multicolumn{1}{l}{\textbf{LogME}} & \multicolumn{1}{l}{\textbf{H-Score}} & \multicolumn{1}{l}{\textbf{TransRate}} & \multicolumn{1}{l}{\textbf{\method}} \\
		\midrule
		Caltech-101         & +        & -       & -        & -          & -        & +            & +                   \\
		Caltech-256         & +        & -       & -        & -          & -        & +            & +                   \\
		CIFAR-10            & +        & -       & -        & -          & -        & +            & +                   \\
		CIFAR-100           & +        & -       & -        & -          & -        & +            & +                   \\
		MNIST               & -        & -       & -        & -          & +        & +            & +                   \\
		FashionMNIST        & +        & -       & -        & -          & -        & +            & +                   \\
		SVHN                & -        & -       & -        & -          & -        & +            & +                   \\
		FlowerPhotos        & +        & -       & -        & -          & -        & +            & +                   \\
		EuroSAT             & +        & -       & -        & -          & +        & +            & +                   \\
		Chest X-Ray         & +        & -       & -        & -          & -        & +            & +                   \\
		VisDA               & +        & -       & -        & -          & +        & +            & +                   \\
		FGVC-Aircraft  		& +        & -       & -        & -          & -        & +            & +                   \\
		CUB-200             & +        & -       & -        & -          & -        & +            & +                   \\
		Cars                & -        & -       & -        & -          & -        & +            & +                   \\
		DTD                 & +        & -       & -        & -          & -        & +            & +                   \\
		Food-101            & +        & -       & -        & -          & -        & +            & +                   \\
		Oxford-IIIT Pet     & +        & -       & -        & -          & -        & +            & +                   \\
		\bottomrule
	\end{tabular}
	}
\end{table}

\subsection{General Applications}
\label{subsec:exp_application}

\paragraph{Applying to self-supervised pre-trained models.}
\method not only measures the transferability between supervised pre-trained models and classification tasks but also can be extended to self-supervised pre-trained models.
We prepare eleven self-supervised pre-trained models already trained on ImageNet provided by Github repositories introduced in Section~\ref{subsec:setup}.
NCE~\cite{nce} and LEEP~\cite{leep} cannot measure the transferability between a self-supervised pre-trained model and a downstream task because they dig out the relationship between target ground truth and predicted label or probability.
Thus, we compare the proposed \method to LogME~\cite{logme}, H-Score~\cite{H-score}, and TransRate~\cite{transrate} in Table~\ref{table:ssl}; \method outperforms competitors in 13 out of 17 classification datasets.

\begin{table}[!t]
	\caption{
	Correlations between fine-tuned accuracies and transferability scores when transferring self-supervised pre-trained models to classification tasks.
	The best result is in bold.
	\method shows the best Kendall correlation consistently in 13 out of 17 cases, outperforming all the competitors.
	}
	\centering
	\label{table:ssl}
	\scalebox{0.8}{
	\begin{tabular}{lcccc}
		\toprule
        \multirow{2}{*}{\textbf{Dataset}}	& \multicolumn{4}{c}{\textbf{Kendall Correlation Coefficient}}					\\
            								& \textbf{LogMe} 	& \textbf{H-Score} & \textbf{TransRate}	& \textbf{\method} 	\\
        \midrule
		Caltech-101     & -0.278			  & 0.183           & 0.000          & \textbf{0.367} \\
		Caltech-256     & -0.561			  & 0.345           & 0.164          & \textbf{0.600} \\
		CIFAR-10        & -0.337          & 0.018           & 0.164          & \textbf{0.624} \\
		CIFAR-100       & -0.561          & 0.673           & 0.236          & \textbf{0.697} \\
		MNIST           & -0.574          & -0.262		    & -0.426         & \textbf{0.246} \\
		FashionMNIST    & -0.330          & -0.220          & -0.330         & \textbf{0.094} \\
		SVHN            & 0.055           & \textbf{0.309}  & -0.345         & 0.075			  \\
		FlowerPhotos    & \textbf{0.389}  & -0.241          & 0.130          & 0.075          \\
		EuroSAT         & -0.073 		  & -0.037		 	& \textbf{0.037} & -0.208         \\
		Chest X-Ray     & -0.037          & -0.037	 		& -0.257         & \textbf{0.175} \\
		VisDA           & 0.550           & 0.055           & \textbf{0.309} & -0.597         \\
		FGVC-Aircraft   & -0.345          & 0.091           & -0.309         & \textbf{0.150} \\
		CUB-200         & -0.112          & -0.018          & 0.091          & \textbf{0.564} \\
		Cars            & 0.000           & \textbf{0.382}  & -0.200         & 0.262          \\
		DTD             & 0.382           & 0.018		    & 0.224          & \textbf{0.455} \\
		Food-101        & -0.411          & -0.018          & -0.018         & \textbf{0.204} \\
		Oxford-IIIT Pet & -0.585          & 0.294           & 0.073          & \textbf{0.807} \\

		\bottomrule
	\end{tabular}
	}
\end{table}

\paragraph{Selecting the best transferring layer.}
Given a ResNet-18 trained on ImageNet, we can easily determine the best transferring layer, which produces the best fine-tuning performance for a target task by measuring transferability.
We compare the correlations of LogME~\cite{logme}, H-Score~\cite{H-score}, and TransRate to select the best transferring layer for fine-tuning.
Theoretically, LogME~\cite{logme} and H-Score~\cite{H-score} cannot deal with layer selection because they rely on features from the ultimate layer of the feature extractor.
However, we extract features from each layer and directly put them into LogME and H-Score algorithms to calculate the transferability.
The size of extracted features is quite large in the early layers, resulting in much higher computational cost for transferability measurement. 
To reduce this cost, we add an avgpool2d layer to the end of transferring layer to reduce the dimension of extracted features.
Table~\ref{table:layer} shows that \method achieves the highest correlation coefficients in 16 out of 17 datasets.
Correlation coefficients of MNIST, SVHN, and Chest X-Ray are much lower than those of the other datasets because their ranges of minimum and maximum accuracies obtained by layer transfer are smaller than 0.5\%, making it challenging to rank these dataset accurately.

\begin{table}[!t]
	\caption{
	Correlations between fine-tuned accuracies and transferability scores when transferring specific layers of pre-trained models.
	The best result is in bold.
	\method shows the best correlations consistently in 16 out of 17 cases.
	}
	\centering
	\label{table:layer}
	\scalebox{0.8}{
	\begin{tabular}{lcccc}
		\toprule
        \multirow{2}{*}{\textbf{Dataset}}	 & \multicolumn{4}{c}{\textbf{Kendall Correlation Coefficient}}					\\
						                     & \textbf{LogMe}	& \textbf{H-Score} & \textbf{TransRate}	& \textbf{\method}  \\
        \midrule
		Caltech-101     & 0.346 & 0.935          & 0.752 & \textbf{0.941} \\
		Caltech-256     & 0.973 & 0.317          & 0.800 & \textbf{0.984} \\
		CIFAR-10        & 0.360 & 0.743          & 0.704 & \textbf{0.746} \\
		CIFAR-100       & 0.460 & 0.774          & 0.774 & \textbf{0.801} \\
		MNIST           & 0.180 & 0.055          & 0.179 & \textbf{0.184} \\
		FashionMNIST    & 0.294 & 0.241          & 0.298 & \textbf{0.308} \\
		SVHN            & 0.087 & 0.106          & 0.040 & \textbf{0.113} \\
		FlowerPhotos    & 0.779 & 0.198          & 0.687 & \textbf{0.792} \\
		EuroSAT         & 0.691 & 0.709          & 0.495 & \textbf{0.735} \\
		Chest X-Ray     & 0.113 & \textbf{0.154} & 0.100 & 0.054          \\
		VisDA           & 0.665 & 0.866          & 0.826 & \textbf{0.904} \\
		FGVC-Aircraft   & 0.791 & 0.302          & 0.595 & \textbf{0.817} \\
		CUB-200         & 0.317 & 0.970          & 0.787 & \textbf{0.954} \\
		Cars            & 0.330 & 0.869          & 0.752 & \textbf{0.882} \\
		DTD             & 0.924 & 0.401          & 0.761 & \textbf{0.931} \\
		Food-101        & 0.409 & 0.948          & 0.869 & \textbf{0.970} \\
		Oxford-IIIT Pet & 0.349 & 0.970          & 0.826 & \textbf{0.990} \\
		\bottomrule
	\end{tabular}
	}
\end{table}

\subsection{Hyperparameter Sensitivity}
\label{subsec:exp_hyp}

\begin{figure}[!t]
\vskip 0.2in
	\begin{center}
 	\scalebox{0.95}{
	\centerline{\includegraphics[width=0.5\columnwidth]{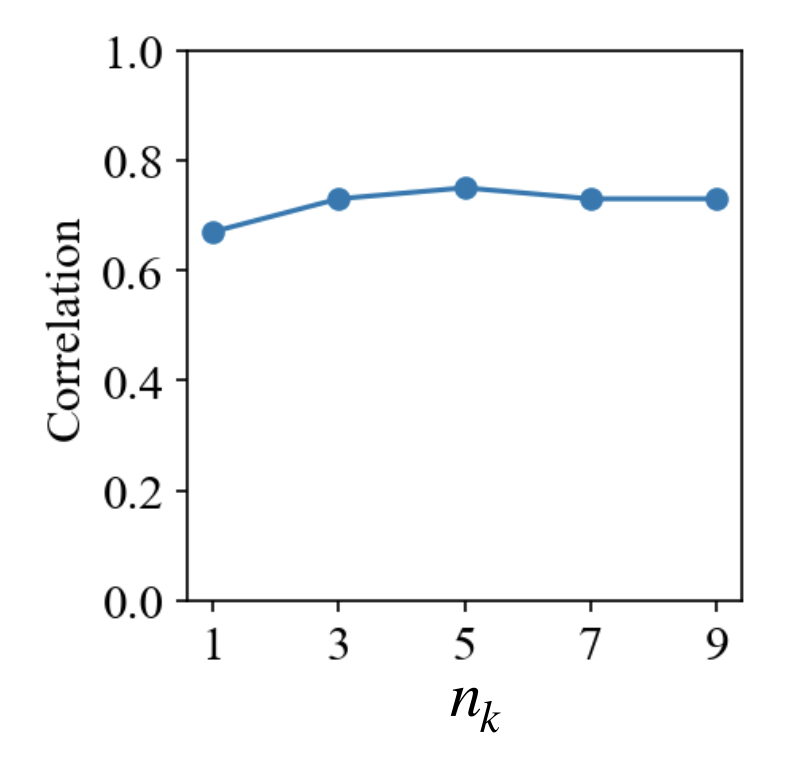}}
	}
	\caption{Sensitivity to the hyperparameter $n_k$ in entropy estimator.}
	\label{fig:sensitivity}
	\end{center}
\vskip -0.2in
\end{figure}

An advantage of \method is that it does not train models
on target datasets because it is time-consuming and sensitive to hyperparameters which may cause different results.
The only hyperparameter in \method is $n_k$ used in the entropy estimator.
In Figure.~\ref{fig:sensitivity}, the correlation between pre-trained models and \method shows consistency.



\section{Conclusion}
\label{sec:Conclusion}
We propose \method, a fast, accurate, and general algorithm to measure the transferability.
\method quantifies the degree of generalization of a pre-trained model to a new task by estimating the intra-class feature variance.
We conduct experiments on seventeen datasets with fifty supervised pre-trained models and eleven self-supervised pre-trained models.
\method selects the top-5 pre-trained architecture in 14 out of 17 cases and selects the best source data in all cases, outperforming its competitors.
We measure the correlation between the fine-tuning performance and transferability, and our proposed \method outperforms competitors in 13 out of 17 cases.

\section*{Acknowledgments}
\label{sec:ack}

This work was supported by Institute of Information \& communications Technology Planning \& Evaluation(IITP) grant funded by the Korea government(MSIT) [No.2020-0-00894, Flexible and Efficient Model Compression Method for Various Applications and Environments], [No.2021-0-01343, Artificial Intelligence Graduate School Program (Seoul National University)], and [NO.2021-0-02068, Artificial Intelligence Innovation Hub (Artificial Intelligence Institute, Seoul National University)].
The Institute of Engineering Research at Seoul National University provided research facilities for this work.
The ICT at Seoul National University provides research facilities for this study.
U Kang is the corresponding author.

{\small
\bibliographystyle{ieee_fullname}
\bibliography{ref}
}

\end{document}